\documentclass{article}
\usepackage{spconf,amsmath,graphicx}
\usepackage{times}
\usepackage{epsfig}
\usepackage{graphicx}
\usepackage{amsmath}
\usepackage{amssymb}
\usepackage{booktabs}
\usepackage{nidanfloat}
\usepackage{color}
\usepackage{multirow}
\usepackage{tikz}
\usepackage{pgfplots}
\usepackage{cite}
\usepackage{caption}
\usepackage{lipsum}
\let\OLDthebibliography\thebibliography
\renewcommand\thebibliography[1]{
  \OLDthebibliography{#1}
  \setlength{\parskip}{0pt}
  \setlength{\itemsep}{0pt plus 0.3ex}
}

\title{Beyond Caption To Narrative: \\ Video Captioning With Multiple Sentences}
\name{Andrew Shin, Katsunori Ohnishi, and Tatsuya Harada}
\address{Grad. School of Information Science and Technology, The University of Tokyo, Japan}
\begin{document}
%
\maketitle
\begin{abstract}
Recent advances in image captioning task have led to increasing interests in video captioning task. However, most works on video captioning are focused on generating single input of aggregated features, which hardly deviates from image captioning process and does not fully take advantage of dynamic contents present in videos. We attempt to generate video captions that convey richer contents by temporally segmenting the video with action localization, generating multiple captions from multiple frames, and connecting them with natural language processing techniques, in order to generate a story-like caption. We show that our proposed method can generate captions that are richer in contents and can compete with state-of-the-art method without explicitly using video-level features as input.
\end{abstract}
\section{Introduction}
\label{sec:intro}

Image captioning task has gained an increasing amount of attention with successful application of  convolutional neural networks (CNN) and recurrent neural networks (RNN), setting new benchmarks that are frequently comparable to human-written captions \cite{Chen, Fang, Karpathy, Vinyals}. Successes in image captioning task have inevitably led to an increasing amount of interests in video captioning task. Previous works dealing with video captioning task have attempted to utilize various type of features that are unique to videos. However, most of them were focused on generating caption from single input vector of aggregated features, which hardly deviates from image captioning process and does not take full advantage of richer contents present in videos. 

Visual elements are hardly stationary in videos, which inevitably leads to a \textit{series} of events, and as such, caption generated from single aggregated input can be insufficient for fully conveying the dynamic contents of videos. It consequently follows that we need to account for each event that is happening in the video with multiple sentences. Segmentation of videos is an intuitive method through which we can generate multiple sentences, provided that we have a reasonable criteria for segmentation. One feasible way is to make use of scene or shot transition detection. However, most such techniques revolve around color histograms and edge features\cite{ShotColor, ShotEdge}. Thus, while effective in highly edited videos such as movie clips, it is difficult to exert an equal amount of efficiency in videos of other domains where scene or shot transition is less frequent. Even in videos with frequent editing, shot or scene transition does not always correspond to transition of events. For example, it is counter-intuitive to segment repeated counter-shot editing of two people talking to each other and refer to it as separate events. We thus attempt a segmentation of videos based on action localization, as change of action can be a strong indicator of a distinct event.
\begin{figure}
\centering
{\includegraphics[width=6cm,height=5.75cm]{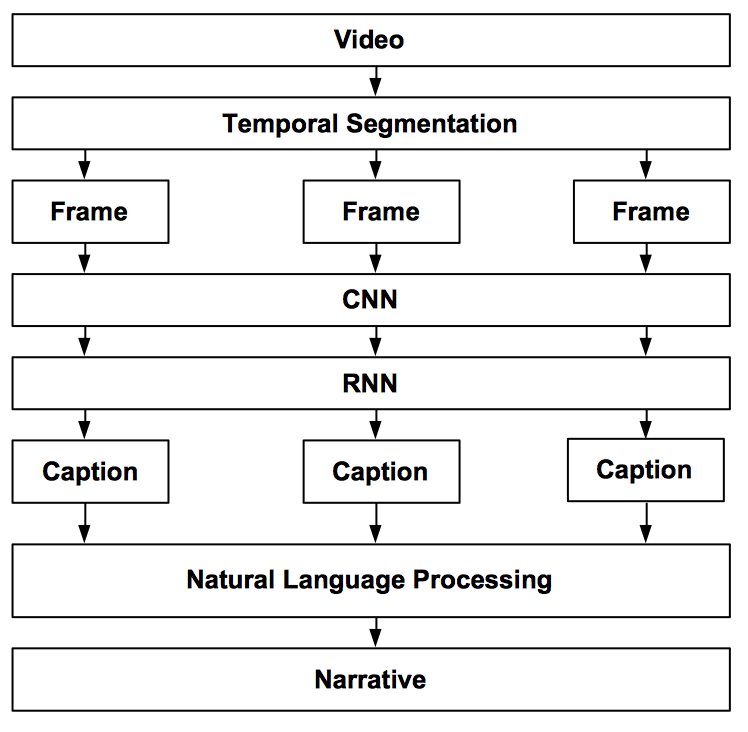}} \\
\caption{Overall workflow of our proposed method}
\label{fig:captions}
\vspace{-3ex}
\end{figure}

Segmentation of videos allows us to work with multiple frames, with which we can generate separate captions following the conventional image captioning procedures. However, these captions are initially independent of each other, bearing no contextual relevance. Thus, simply concatenating the sentences leads to a highly awkward and artificial-looking passage. We employ a series of simple natural language processing techniques to connect the sentences smoothly, enabling a generation of more natural, human-like captions, resembling the characteristics of \textit{narratives}. Fig. 1 shows the overall workflow of our proposed method. We demonstrate through experiments that our method, even without explicitly using video-level features as input, can generate captions conveying richer contents, with comparable performance to current state-of-the-art on evaluation metrics.

\section{Related Work}
\label{sec:format}

Human action recognition in videos has been actively studied. The typical pipeline to obtain video features is to extract local features and aggregate them. Rich descriptors, such as HOG \cite{dalal2005histograms}, HOF \cite{lucas1981iterative}, and MBH \cite{dalal2006human}, and aggregating these local features with Fisher Vector \cite{perronnin2007fisher} have demonstrated successful performances. Extracting local features along improved Dense Trajectory (iDT) \cite{wang2013action}, which compensates for camera motion, can obtain impressive results in action recognition. Recently, deep CNN features have been employed in a variety of fields and demonstrated successes \cite{Cai,Yang}. Yet, while the state-of-the-art method \cite{wang2015action} in action recognition uses CNN, it also depends on iDT. Spatio-temporal localization has also been attempted using such representations of videos. Most of the state-of-the-art works on action localization use temporal sliding window approach \cite{oneata2013action, weinzaepfel2015learning}. In our case, it suffices to temporally segment the videos based on actions, and we thus localize the actions in temporal axis only. 

Works on video captioning have attempted to directly apply a variety of features uniquely present in videos to caption generation. Pan et al. \cite{Pan} introduced LSTM-E, which simultaneously takes learning of LSTM and visual-semantic embedding into account. Shetty et al.\cite{Workshop}, winner of Describing and Understanding Video \& The Large Scale Movie Description Challenge (LSMDC) at ICCV 2015, extracted features at three levels. At video level, they extract Dense Trajectory (DT)\cite{DT}, HOG, HOF, MBH (\textit{x} and \textit{y}) and generate five histograms of 1000 dimensions. Then, the keyframe features from the middle frame of the video are extracted via VGG-16, VGG-19\cite{VGG}, and GoogLeNet\cite{Google}. Finally, content features are extracted by training a SVM on MS COCO 80-object classification. While competitive, a few potential drawbacks can be pointed out. First, the video representation becomes high-dimensional. Also, it is questionable whether middle frame is always representative of the entire video. It may be true for short videos, but as the duration of video gets longer, its representativeness is inevitably prone to decline. Lastly, in spite of the relatively complex overall pipeline, its caption generation process hardly differs from image captioning in that it ends up with a single representation, which may not reflect rich contents unique in videos.

In terms of the story-like characteristic of the outcome, Zhu et al.\cite{ICCVBook} are more intimate to our motivation. By aligning scenes with subtitles, and subtitles with passages in the books, they were able to generate story-like description of the scenes. This, however, has an obvious drawback that both books and movies should exist, along with subtitles. Alternatively, one may be able to rely on story-like, human-written ground truth captions for images or videos, but such data are not yet sufficiently available in the research community, and are expensive to collect. Since our work treats the frames from each segment independently, we do not need to rely on additional textual data to generate a story-like caption.

\section{Proposed Method}
We first describe our mechanism of temporal segmentation based on action localization in Sec. 3.1. Middle frames are extracted from each resulting segment, and captions are generated for each frame. These captions are connected via two steps of natural language processing, namely coreference resolution and connective word generation, which we explain in Sec. 3.2 and Sec. 3.3, respectively.

\subsection{Temporal Segmentation with Action Localization}
\label{sec:pagestyle}
\begin{figure}
\begin{tikzpicture}
\centering
\pgfplotsset{
}
\begin{axis}[
      width=0.5\textwidth,
      height=0.25\textwidth,
ylabel near ticks, yticklabel pos=left,
  ymin=0, ymax=5,
  xmin=-1.1,xmax=0,
  xlabel=Score Threshold,
  ylabel=Avg. \# of Segments,
    ymajorgrids=true,
    grid style=dashed,
]
\addlegendentry{Montreal}
\addplot[ultra thick,mark=x,blue]
  coordinates{
(-0.1,1.00121212121)
(-0.2,1.32175318118)
(-0.3,	1.67986265401)
(-0.4,1.9852180937)
(-0.5,	2.47939393939)
(-0.6,2.8580658187)
(-0.7,	3.2068617558)
(-0.8,3.50252270434)
(-0.9,	3.73057517659)
(-1.0,	3.91868442292)
};
\addplot[ultra thick,mark=x,red]
  coordinates{
(-0.1,	0.142553191489)
(-0.2,0.204255319149)
(-0.3,	0.285714285714)
(-0.4,0.392249240122)
(-0.5,	0.525987841945)
(-0.6,0.679635258359)
(-0.7,	0.860182370821)
(-0.8,1.05699088146)
(-0.9,	1.2509118541)
(-1.0,1.4490881459)
}; 
\addlegendentry{MPII}
\end{axis}
\end{tikzpicture}
\caption{Average number of segments per video depending on score threshold. Note that the number of segments was set to zero in case no action was localized.}
\vspace{-3mm}
\end{figure}
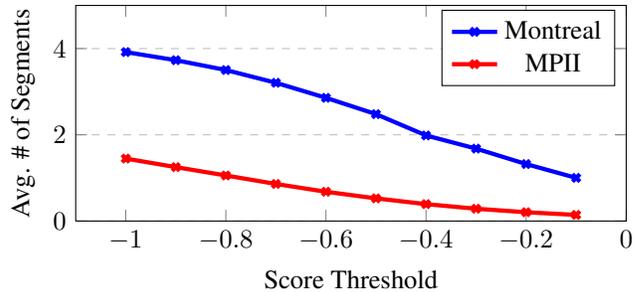

It has been argued that features extracted from pre-trained CNN \cite{xu2015discriminative, jain201515} or color channels\cite{clinchant2008trans} are more appropriate for capturing contextual information rather than localizing actions precisely. We thus employ a sliding window method and motion features only for temporal localization, following the winner's approach in the temporal action detection task of THUMOS 2014 challenge \cite{oneata2014lear}. In order to extract motion features, we use improved Dense Trajectories (iDT) \cite{wang2013action}.

Since datasets we work on (Montreal\cite{Montreal} and MPII\cite{MPII}) do not contain time information, it is difficult to learn the actions directly from these datasets. However, since our goal is simply to segment the videos, predicting the action class is not a prerequisite. We thus rely on UCF 101\cite{soomro2012ucf101} to learn the actions. Although it cannot be said that action classes in UCF 101 cover all actions (and verbs in ground truths) in Montreal or MPII, it is still possible to obtain class scores using a classifier trained on UCF 101. Since our goal is to temporally segment the video based on action change, rather than specifying the action, we apply the classifier trained on UCF 101 training data to test data of our target datasets.

Following \cite{oneata2013action}, we employ a temporal sliding window approach and non-maximum suppression for temporal action localization. We first use temporal sliding windows of length 30, 60, 90, and 120 frames, and slide the windows in steps of 30 frames. We then re-score the detected windows with non-maximum suppression to remove overlapping. Note that if we were to use all detected windows, then minor movements that do not quite constitute a clear action will be localized as well, leading to unnecessary, redundant generation of captions. We thus set a score threshold so that windows with score below it are not considered to contain any action. We segment the videos so that each segment consists of adjacent windows of the same action class.

Fig. 2 shows the average number of segments per video depending on the score threshold. Although we want to extract multiple frames from each video, too many frames will result in excessively wordy and repetitive caption. We thus set our score threshold as -0.5 for Montreal dataset, which is just over 2 segments per video. MPII dataset is relatively shorter in length compared to Montreal, and all of our thresholds did not exceed 2 segments per video. We choose threshold of -1.0 for MPII as it closely approaches 2 frames per video. Note that, since we assumed the number of segments to be zero when no action was detected, the actual number of segments is a bit higher than shown on the graph. In case no action was detected in the entire video, we simply used the middle frame.

\subsection{Backward Coreference Resolution}
\label{sec:typestyle}
As was discussed in Sec. 1, generated captions from multiple frames of a video are initially independent of each other with no contextual relevance. As the first step to smoothly connect them, we first apply \textit{backward} coreference resolution. In usual coreference resolution, given a passage\\
\centerline{\textit{A man is with a plate. He is sitting with it.}}
anaphors are resolved to their corresponding antecedents so that `\textit{he}' is linked to a `\textit{a man},' and `\textit{it}' to `\textit{a plate}.' In our case, on the contrary, there are no anaphora but repeated appearances of to-be-antecedent noun phrases. Thus, given a passage\\
\centerline{\textit{A man is with a plate. A man is sitting with a plate.}}
we want to link the coreferences, and convert the later ones to appropriate pronouns.

Such task is rare even in natural language processing field, and to our knowledge, there exists no tool to specifically perform this task. We implement the task by using gender annotator from Stanford CoreNLP\cite{Stanford}, which assigns likely gender to tokens, in order to deal with singular human subjects. Then we run coreference resolution for non-singular or non-human references, with part-of-speech tagger to deal with plurals (NNS/NNPS), and with lemmatization, in order to avoid conflicts with tenses \cite{stack}(`\textit{watches}' for noun plural vs. `\textit{watches}' as a third person verb), aided by WordNet \cite{WordNet}.

\subsection{Connective Word Generation}
\label{sec:majhead}
Another key point to generating human-like `narrative' from multiple sentences is to find appropriate transition, or connective words, such as `\textit{then}.' One way to accomplish this may be to train a classifier for the connective word based on vectorized adjacent sentences. This, however, is not very practical as it is hard to determine the number of classes, and collecting training data is expensive, requiring an extensive usage of a parser. We thus tackle the task in an unsupervised way.

\begin{figure}
\centering
{\includegraphics[width=8cm,height=5cm]{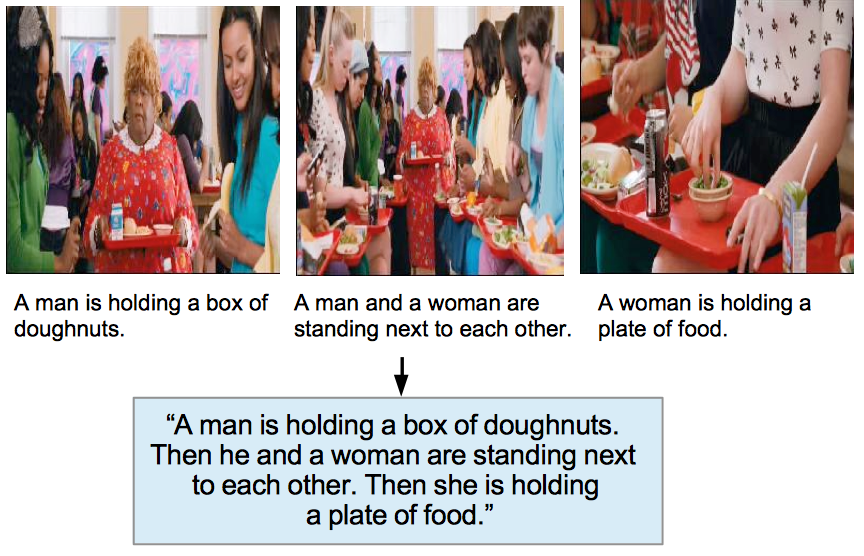}} \\
\begin{tabular}{ c|l }
\hline \multirow{2}{*}{\bf Ground Truth} & With the table gone, all of the girls sit in\\
& their same seats, their trays on their laps.\\ \hline
\multirow{2}{*}{\bf Mid-frame} & A man and a woman standing next to\\
& each other. \\ \hline
\end{tabular}
\caption{Example of generated caption on a test example from Montreal dataset, along with ground truth and caption from middle frame}
\label{fig:captions}
\vspace{-0ex}
\end{figure}

 \begin{table}[t!]
\small
\begin{center}
\caption{Performances of our model and baseline on each dataset}
\begin{tabular}{c|c|cccccccc}
\hline \bf Dataset & \bf Model & \bf BLEU-4 & \bf CIDEr & \bf METEOR\\ \hline 
\multirow{2}{*}{Montreal} & Mid-frame & .003 & .070 & .042 \\ 
& Ours & \bf .004 & \bf .089 & \bf .047 \\ \hline 
\multirow{2}{*}{MPII} & Mid-frame & .009 & .065 & .046 \\ 
& Ours & \bf .013 & \bf .075 & \bf .048 \\ \hline  
MS Video & Mid-frame & .043 & .148 & .107 \\ 
(subset) & Ours &  \bf .063 & \bf .177 &  .104 \\ \hline
\end{tabular}
\label{table:scores}
\end{center}
\vspace{-2ex}
\end{table}

\begin{table}[t!]
\small
\begin{center}
\caption{Performances of each model on automatic evaluation metrics for LSMDC. Note that scores for \cite{Workshop} are the best scores as reported in the corresponding paper.}
\begin{tabular}{@{ \ }c@{ \ \ }|@{ \ \ }c@{ \ \ }c@{ \ \ }c@{ \ \ }c@{ \ \ }c@{ \ }}
\hline \bf Model & \bf Avg. Length & \bf BLEU-4 & \bf CIDEr  &\bf METEOR \\ \hline
\cite{Workshop} & 5.33 & \bf .006 & .092 & .058\\ \hline
Mid-frame & 9.78 & .004 & .068 & .044\\ \hline
Ours  & \bf 19.34 & \bf .006 & .085 & .048 \\
\hline
\end{tabular}
\label{table:human}
\end{center}
\vspace{-5ex}
\end{table}

We collected 500 instances of adjacent sentences from tagged version of Wikicorpus\cite{Wiki}, in which the second sentence starts with a connective word. Specifically, we ran a CKY parser for PCFG trained on Penn Treebank\cite{Penn}, and retained the cases in which the second sentence starts with a single word of adjective (JJ) or adverb (RB) tag as a part of an adjective phrase (AJP), followed directly by full sentence (S) consisting of noun phrase (NP) and verb phrase (VP). Collected instances, excluding the connective word, are converted to 300-dimensional vector by Sentence2Vec\cite{Sentence2Vec}. The rest is a simple matching of vectorizing the adjacent sentences of generated captions in the same way, finding the closest instance by L2 distance, and inserting the connective word present in the instance to the captions.

\section{Experiment}
\label{ssec:subhead}

\begin{figure}
\centering
{\includegraphics[width=8cm,height=4cm]{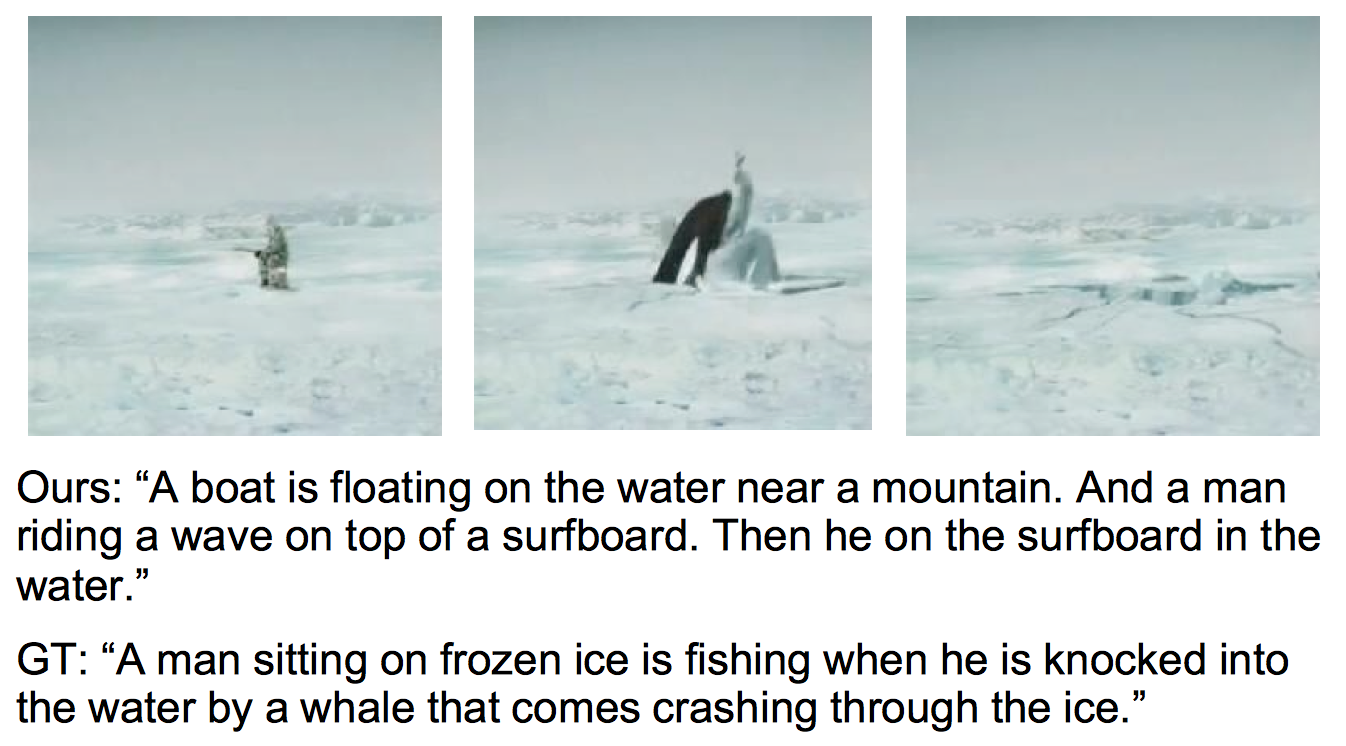}} \\
\caption{Example of generated caption on non-movie video clip from MS Video Description Corpus with ground truth. Note that scene or shot transition rarely occurs.}
\label{fig:captions}
\vspace{-2ex}
\end{figure}

We first apply our method to test split of Montreal dataset and MPII dataset, which amount to 11,529 video clips. Note that train and validation splits of both datasets were never used throughout our experiment. As our first baseline, we simply generate captions from CNN features of single middle frame of the videos. In order to obtain motion features, we first extract HOG, HOF, and MBH along iDT and encode obtained local descriptors by Fisher Vector (FV) \cite{perronnin2007fisher}. We reduce the dimension of each descriptor by a factor of 2 with PCA. We then assign descriptors to 256 components of FV and obtain video features. Finally, we apply power and L2-normalization to video features. As classifier, we employ one-vs-rest linear SVM and set $C=100$. 

After segmentation according to action localization is complete, we extract the middle frame from each segment, and extract 4096-dimensional CNN features from the second fully-connected layer (fc7) of VGG trained on ImageNet \cite{ImageNet} using Caffe framework \cite{Caffe}. These features are passed onto LSTM \cite{LSTM} trained on MS COCO dataset \cite{MSCOCO} with beam size 1, so that descriptive caption for each frame can be generated. Once the captions are generated, the remaining step is simply to connect them smoothly according to the method specified in Sec. 3.2 and Sec. 3.3

Note that video-level features are used only for the sake of temporal segmentation, and the inputs for captioning are simply CNN features from multiple frames. Neither were there any need for additional textual data accompanying the videos.

Table 1 shows our method's performance on automatic evaluation metrics \cite{BLEU, Meteor, Cider} for each dataset along with baseline. Our method clearly outperforms the baseline across all datasets and all metrics. Fig. 3 shows an example of generated story-like caption, along with ground truth and single-frame captions. We also compare our method to one of the state-of-the-art methods in video captioning, the winner of LSMDC challenge\cite{Workshop}, as shown in Table 2. It should be first noted that, in our case, the average length actually refers to the average length of the entire caption consisting of multiple sentences. It is thus natural that ours is about 4 times longer than the compared method, consequently containing more contents. The question is whether having more contents implies more reflection of the original contents present in the video, or simply more noise and less precision. Results on the metrics show that our method is comparable to the state-of-the-art, in spite of the increased lengths and risk for noises. Although we were not able to outperform the state-of-the-art method, this shows that our longer caption reflects the contents of the video with roughly the same ratio as the state-of-the-art method, thus reflecting more of the video contents.

In order to examine how our method performs on video clips of non-movie domain, we applied our method to a subset (138 videos) of Youtube videos listed in Microsoft Video Description Corpus\cite{MSVideo}. Since these videos are usually minutes long, we set our score threshold higher at -0.1. Note that ground truth captions are written only on sub-interval of the videos in this dataset, so direct comparison cannot be made. We thus extracted our captions from sub-interval specified in the corpus and compared it to 5 ground truths per video. As shown in Table 1, we were able to outperform single frame approach in most metrics, in spite of longer captions generated. This confirms that our method based on action localization can be applied to videos in which shot or scene transitions occur much less frequently than movie clips. Fig. 4 presents an example caption on one of the videos from MS Video Description Corpus along with ground truth.

\section{Conclusion \& Future Work}
We proposed a method that generates a story-like video caption from multiple frames by temporal segmentation using action localization, and natural language processing techniques. We demonstrated that our method outperforms single-frame method, while being comparable to current state-of-the-art method, even with longer captions that contain more contents. This was made possible even without explicit usage of video-level features as input for captioning.

Working on the task, we have come across a few concrete ideas that can potentially improve the result. First, since training an action classifier from an existing dataset limits the number of action classes, it may be a better idea to localize the actions in an unsupervised manner, so that actions not present in the action classes of dataset can also be segmented.

Another potential improvement can be made with face identification. With current pipeline, it is impossible to know whether `\textit{a man}' from two different frames refers to the same person or not. In fact, there were cases in which a new character appears, yet our pipeline treats the new character as the the original character who already appeared. Adding face identification to our pipeline can generate a more robust caption, in which a new character is referred to as `\textit{another man}' or such.

Finally, applying our method to recently published video annotation datasets \cite{MSR} will also be an interesting challenge.
\label{sec:print}

\small
\bibliographystyle{IEEEbib}
\bibliography{/Users/andrewshin/Downloads/ICIP_tex/refs}

\begin{thebibliography}{10}

\bibitem{Chen}
X.~Chen and C.~Zitnick,
\newblock ``Mind's eye: A recurrent visual representation for image caption
  generation,''
\newblock in {\em CVPR}, 2015.

\bibitem{Fang}
H.~Fang, S.~Gupta, F.~Iandola, R.~Srivastava, L.~Deng, P.~Dollar, J.~Gao,
  X.~He, M.~Mitchell, J.~Platt, C.~Zitnick, and G.~Zweig,
\newblock ``From captions to visual concepts and back,''
\newblock in {\em CVPR}, 2015.

\bibitem{Karpathy}
A.~Karpathy and L.~Fei-Fei,
\newblock ``Deep visual-semantic alignments for generating image
  descriptions,''
\newblock in {\em CVPR}, 2015.

\bibitem{Vinyals}
O.~Vinyals, A.~Toshev, S.~Bengio, and D.~Erhan,
\newblock ``Show and tell: A neural image caption generator,''
\newblock in {\em CVPR}, 2015.

\bibitem{ShotColor}
N.~Janwe and K.~Bhoyar,
\newblock ``Video shot boundary detection based on jnd color histogram,''
\newblock in {\em ICIIP}, 2013.

\bibitem{ShotEdge}
R.~Zabih, J.~Miller, and K.~Mai,
\newblock ``A feature-based algorithm for detecting and classifying scene
  breaks,''
\newblock in {\em ACM MM}, 1995.

\bibitem{dalal2005histograms}
N.~Dalal and B.~Triggs,
\newblock ``Histograms of oriented gradients for human detection,''
\newblock in {\em CVPR}, 2005.

\bibitem{lucas1981iterative}
B.~Lucas and T.~Kanade,
\newblock ``An iterative image registration technique with an application to
  stereo vision.,''
\newblock {\em IJCAI}, vol. 81, pp. 674--679, 1981.

\bibitem{dalal2006human}
N.~Dalal, B.~Triggs, and C.~Schmid,
\newblock ``Human detection using oriented histograms of flow and appearance,''
\newblock in {\em ECCV}, 2006.

\bibitem{perronnin2007fisher}
F.~Perronnin and C.~Dance,
\newblock ``Fisher kernels on visual vocabularies for image categorization,''
\newblock in {\em CVPR}, 2007.

\bibitem{wang2013action}
H.~Wang and C.~Schmid,
\newblock ``Action recognition with improved trajectories,''
\newblock in {\em ICCV}, 2013.

\bibitem{Cai}
J.~Cai, M.~Merler, S.~Pankanti, and Q.~Tian,
\newblock ``Heterogeneous semantic level features fusion for action
  recognition,''
\newblock in {\em ICMR}, 2015.

\bibitem{Yang}
Z.~Xu, Y.~Yang, and A.~Hauptmann,
\newblock ``A discriminative cnn video representation for event detection,''
\newblock in {\em CVPR}, 2015.

\bibitem{wang2015action}
L.~Wang, Y.~Qiao, and X.~Tang,
\newblock ``Action recognition with trajectory-pooled deep-convolutional
  descriptors,''
\newblock in {\em CVPR}, 2015.

\bibitem{oneata2013action}
D.~Oneata, J.~Verbeek, and C.~Schmid,
\newblock ``Action and event recognition with fisher vectors on a compact
  feature set,''
\newblock in {\em ICCV}, 2013.

\bibitem{weinzaepfel2015learning}
P.~Weinzaepfel, Z.~Harchaoui, and C.~Schmid,
\newblock ``Learning to track for spatio-temporal action localization,''
\newblock in {\em ICCV}, 2015.

\bibitem{Pan}
Y.~Pan, T.~Mei, T.~Yao, H.~Li, and Y.~Rui,
\newblock ``Jointly modeling embedding and translation to bridge video and
  language,''
\newblock in {\em CVPR}, 2016.

\bibitem{Workshop}
R.~Shetty and J.~Laaksonen,
\newblock ``Video captioning with recurrent networks based on frame- and
  video-level features and visual content classification,''
\newblock in {\em ICCV Workshop on LSMDC}, 2015.

\bibitem{DT}
H.~Wang, A.~Klaser, C.~Schmid, and C-L. Lin,
\newblock ``Action recognition by dense trajectories,''
\newblock in {\em CVPR}, 2011.

\bibitem{VGG}
K.~Simonyan and A.~Zisserman,
\newblock ``Very deep convolutional networks for large-scale image
  recognition,''
\newblock in {\em ICLR}, 2015.

\bibitem{Google}
C.~Szegedy, W.~Liu, Y.~Jia, P.~Sermanet, S.~Reed, D.~Anguelov, D.~Erhan,
  V.~Vanhoucke, and A.~Rabinovich,
\newblock ``Going deeper with convolutions,''
\newblock in {\em ILSVRC}, 2014.

\bibitem{ICCVBook}
Y.~Zhu, R.~Kiros, R.~Zemel, R.~Salakhutdinov, R.~Urtasun, A.~Torralba, and
  S.~Fidler,
\newblock ``Aligning books and movies: Towards story-like visual explanations
  by watching movies and reading books,''
\newblock in {\em ICCV}, 2015.

\bibitem{xu2015discriminative}
Z.~Xu, Y.~Yang, and A.~Hauptmann,
\newblock ``A discriminative {CNN} video representation for event detection,''
\newblock in {\em CVPR}, 2015.

\bibitem{jain201515}
M.~Jain, J.~van Gemert, and C.~Snoek,
\newblock ``What do 15,000 object categories tell us about classifying and
  localizing actions?,''
\newblock in {\em CVPR}, 2015.

\bibitem{clinchant2008trans}
S.~Clinchant, Renders J, and G.~Csurka,
\newblock ``Trans-media pseudo-relevance feedback methods in multimedia
  retrieval,''
\newblock in {\em Advances in Multilingual and Multimodal Information
  Retrieval}. 2008.

\bibitem{oneata2014lear}
D.~Oneata, J.~Verbeek, and C.~Schmid,
\newblock ``The lear submission at thumos 2014,''
\newblock in {\em ECCV workshop on THUMOS Challenge}, 2014.

\bibitem{Montreal}
A.~Torabi, C.~Pal, H.~Larochelle, and A.~Courville,
\newblock ``Using descriptive video services to create a large data source for
  video annotation research,''
\newblock in {\em arXiv:1503.01070}, 2015.

\bibitem{MPII}
A.~Rohrbach, M.~Rohrbach, N.~Tandon, and B.~Schiele,
\newblock ``A dataset for movie description,''
\newblock in {\em CVPR}, 2015.

\bibitem{soomro2012ucf101}
K.~Soomro, A.~Zamir, and M.~Shah,
\newblock ``Ucf101: A dataset of 101 human actions classes from videos in the
  wild,''
\newblock {\em arXiv:1212.0402}, 2012.

\bibitem{Stanford}
C.~Manning, M.~Surdeanu, J.~Bauer, J.~Finkel, S.~Bethard, and D.~McClosky,
\newblock ``The stanford corenlp natural language processing toolkit,''
\newblock in {\em ACL}, 2014.

\bibitem{stack}
Gabor Angeli,
\newblock ``pronoun resolution backwards,''
  http://stackoverflow.com/questions/34628224/pronoun-resolution-backwards.

\bibitem{WordNet}
G.~Miller,
\newblock ``Wordnet: A lexical database for english,''
\newblock {\em Communications of the ACM}, vol. 38, pp. 39--41, 1995.

\bibitem{Wiki}
S.~Reese, G.~Boleda, M.~Cuadros, L.~Padro, and G.~Rigau,
\newblock ``Wikicorpus: A word-sense disambiguated multilingual corpus,''
\newblock in {\em LREC}, 2010.

\bibitem{Penn}
M.~Marcus, B.~Santorini, and M.~Marcinkiewicz,
\newblock ``Building a large annotated corpus of english: The penn treebank,''
\newblock {\em Computational Linguistics}, vol. 19, pp. 313--330, 1993.

\bibitem{Sentence2Vec}
Q.~Le and T.~Mikolov,
\newblock ``Distributed representations of sentences and documents,''
\newblock in {\em ICML}, 2014.

\bibitem{ImageNet}
J.~Deng, W.~Dong, R.~Socher, L.-J. Li, K.~Li, and L.~Fei-Fei,
\newblock ``Imagenet: A large-scale hierarchical image database,''
\newblock in {\em CVPR}, 2009.

\bibitem{Caffe}
Y.~Jia, E.~Shelhamer, J.~Donahue, S.~Karayev, J.~Long, R.~Girshick,
  S.~Guadarrama, and T.~Darrell,
\newblock ``Caffe: Convolutional architecture for fast feature embedding,''
\newblock in {\em ACM MM}, 2014.

\bibitem{LSTM}
S.~Hochreiter and J.~Schmidhuber,
\newblock ``Long short-term memory,''
\newblock {\em Neural Computation}, vol. 9, pp. 1735--1780, 1997.

\bibitem{MSCOCO}
T.~Lin, M.~Maire, S.~Belongie, J.~Hays, P.~Perona, D.~Ramanan, P.~Dollar, and
  C.~Lawrence Zitnick,
\newblock ``Microsoft coco: Common objects in context,''
\newblock in {\em ECCV}, 2014.

\bibitem{BLEU}
K.~Papineni, S.~Roukos, T.~Ward, and W.~Zhu,
\newblock ``Bleu: A method for automatic evaluation of machine translation,''
\newblock in {\em ICLR}, 2015.

\bibitem{Meteor}
M.~Denkowski and A.~Lavie,
\newblock ``Meteor universal: Language specific translation evaluation for any
  target language,''
\newblock in {\em EACL Workshop on Statistical Machine Translation}, 2014.

\bibitem{Cider}
R.~Vedantam, C.~Lawrence Zitnick, and D.~Parikh,
\newblock ``Cider: Consensus-based image description evaluation,''
\newblock in {\em CVPR}, 2015.

\bibitem{MSVideo}
D.~Chen and W.~Dolan,
\newblock ``Collecting highly parallel data for paraphrase evaluation,''
\newblock in {\em ACL}, 2011.

\bibitem{MSR}
J.~Xu, T.~Mei, T.~Yao, and Y.~Rui,
\newblock ``Msr-vtt: A large video description dataset for bridging video and
  language,''
\newblock in {\em CVPR}, 2016.

\end{thebibliography}

\end{document}